\documentclass[letterpaper, 10 pt, conference]{ieeeconf}  

\IEEEoverridecommandlockouts                              

\usepackage{cite}
\usepackage{textcomp}
\usepackage{amsmath,amsfonts}
\usepackage{algorithm}
\usepackage{algpseudocode}
\usepackage{array}
\usepackage[caption=false,font=normalsize,labelfont=sf,textfont=sf]{subfig}
\usepackage{stfloats}
\usepackage[dvipsnames]{xcolor}
\usepackage{cite}
\usepackage{url}

\usepackage{multirow}
\usepackage{verbatim}
\usepackage{graphicx}

\title{\LARGE \bf
A Hierarchical Deep Reinforcement Learning Framework for Traffic Signal Control with  Predictable Cycle Planning
}

\author{Hankang Gu, Yuli Zhang, Chengming Wang, Ruiyuan Jiang, Ziheng Qiao, Pengfei Fan, Dongyao Jia
\thanks{The authors are with the School of Advanced Technology, Xi'an Jiaotong-Liverpool University, Suzhou 215123, China}
\thanks{This work was supported in part by National Natural Science Foundation of China (No. 62372384) , in part by Suzhou Science and Technology Development Planning Programme (ZXL2024342) and in part by XJTLU Postgraduate Research Scholarship (FOSA2106053)}
}

\begin{document}

\maketitle
\thispagestyle{empty}
\pagestyle{empty}
\begin{abstract}
Deep reinforcement learning (DRL) has become a popular approach in traffic signal control (TSC) due to its ability to learn adaptive policies from complex traffic environments. Within DRL-based TSC methods, two primary control paradigms are ``choose phase" and ``switch" strategies. Although the agent in the choose phase paradigm selects the next active phase adaptively, this paradigm may result in unexpected phase sequences for drivers, disrupting their anticipation and potentially compromising safety at intersections. Meanwhile, the switch paradigm allows the agent to decide whether to switch to the next predefined phase or extend the current phase. While this structure maintains a more predictable order, it can lead to unfair and inefficient phase allocations, as certain movements may be extended disproportionately while others are neglected.
In this paper, we propose a DRL model, named Deep Hierarchical Cycle Planner (DHCP),  to allocate the traffic signal cycle duration hierarchically. A high-level agent first determines the split of the total cycle time between the North-South (NS) and East-West (EW) directions based on the overall traffic state. Then, a low-level agent further divides the allocated duration within each major direction between straight and left-turn movements, enabling more flexible durations for the two movements. We test our model on both real and synthetic road networks, along with multiple sets of real and synthetic traffic flows. Empirical results show our model achieves the best performance over all datasets against baselines. 

\end{abstract}
\section{Introduction}
Traffic congestion remains one of the most pressing challenges in urban transportation systems. As urbanization accelerates and vehicle ownership increases, the stream on road infrastructure grows, leading to longer travel times, increased fuel consumption, and heightened environmental pollution\cite{hussain2023investigating,geng2016environmental,brilon2013experiences}. Conventional traffic signal control (TSC) methods often struggle to adapt to dynamic traffic patterns, highlighting the urgent need for more intelligent and adaptive solutions\cite{wei2019survey}.

In recent years, model-free deep reinforcement learning (DRL) methods have demonstrated the potential in addressing the limitations of conventional TSC methods\cite{haydari2020deep}. DRL enables traffic controllers to learn optimal policies through interactions with the environment, adapting to real-time traffic conditions and improving overall traffic flow. Both independent and cooperative deep reinforcement learning (DRL) agents have been investigated for deployment across networks of various scales and under diverse traffic flow conditions\cite{wu2021efficient,Liu2025MATLIT,gu2024large,hankang2025Communication,jing2022HALight}. Independent DRL agents suffer from the non-stationary problem, which leads to poor convergence\cite{tan1993multi}. The information sharing technique is then investigated to improve the performance of the agent. PressLight \cite{wei2019presslight} considers the difference between the number of vehicles on downstream lanes and that on upstream lanes. The agents in CoLight share information of local observation via graph attention layers\cite{wei2019colight}. 
Off-policy Nash deep Q-network (OPNDQN) was proposed to search Nash equilibrium among neighboring intersections via a fictitious play mechanism\cite{zhang2023large}. Hierarchical Reinforcement Learning (HRL) methods have also been explored for coordinating multiple agents by defining precise objectives \cite{yang2018hierarchical, vezhnevets2017feudal, xu2021hierarchically}. In HRL, a high-level agent, often referred to as the manager, generates sub-goals for each low-level agent, or worker. The low-level agents then optimize their policies to achieve these sub-goals while pursuing their local objectives.

Among the formulations of agents in existing works, two common choices of action space for agents are ``choose phase" and ``switch"\cite{wei2019survey}. The formulation of ``choose phase" setting allows the agent to select phases freely and dynamically, which creates a significant gap between algorithmic design and real-world deployment\cite{wu2021efficient,Liu2025MATLIT,gu2024large,hankang2025Communication,jing2022HALight}. In practice, traffic signal control systems often follow a round-robin cycle or a predetermined phase sequence to ensure predictability and coordination across intersections\cite{hunt1982scoot}. The flexible, non-cyclic phase selection used in DRL can lead to variable and unpredictable cycle lengths, which may confuse drivers and increase the likelihood of collisions. To address this issue, some methods adopt the formulation of ``switch" setting where the phase sequence is fixed and the agent only decides whether to switch to the next phase \cite{wei2018intellilight,van2016coordinated,mannion2016experimental}. One classic phase sequence is North-South Straight (NSS), North-South Left Turn (NSL), East-West Straight (EWS) and East-West Left Turn (EWL). However, many of these approaches result in variable cycle lengths, which diverge from conventional traffic engineering practices that typically assume fixed or predictable cycle timings. This inconsistency can complicate integration with existing infrastructure and hinder coordination with adjacent intersections, potentially limiting the practical applicability of such strategies in real-world deployments.

To address the aforementioned issues, we adopt the idea of HDRL and apply the deep deterministic policy gradient (DDPG) algorithm\cite{lillicrap2015continuous} to allocate the durations of individual signal phases constrained with fixed cycle lengths and a predetermined phase sequence. In our approach, a high-level agent observes the overall traffic state at a given intersection and determines the allocation proportions between two primary traffic directions: north–south (NS) and east–west (EW). Subsequently, a low-level agent observes the traffic state corresponding to a specific direction, along with the allocated duration, and further determines the proportion of time assigned to different signal phases, such as through (straight) and left-turn movements, for that direction. Our main contributions are listed as follows:
\begin{itemize}
    \item We propose a hierarchical DRL model for traffic signal cycle planning, which incorporates two types of agents. The high-level agent operates from a broader perspective to allocate the total cycle duration between two traffic directions. The low-level agent then focuses on optimizing the distribution of phase durations, such as straight and left-turn movements, within each individual direction. 
    \item We test our model on three traffic networks with both real and synthetic traffic flow. Empirical results demonstrates the superiority of our model over all baselines under both real and synthetic scenarios.
\end{itemize}
\section{Preliminary}
\subsection{Traffic Signal Control}
\begin{figure}
    \centering
    \includegraphics[width=0.5\textwidth]{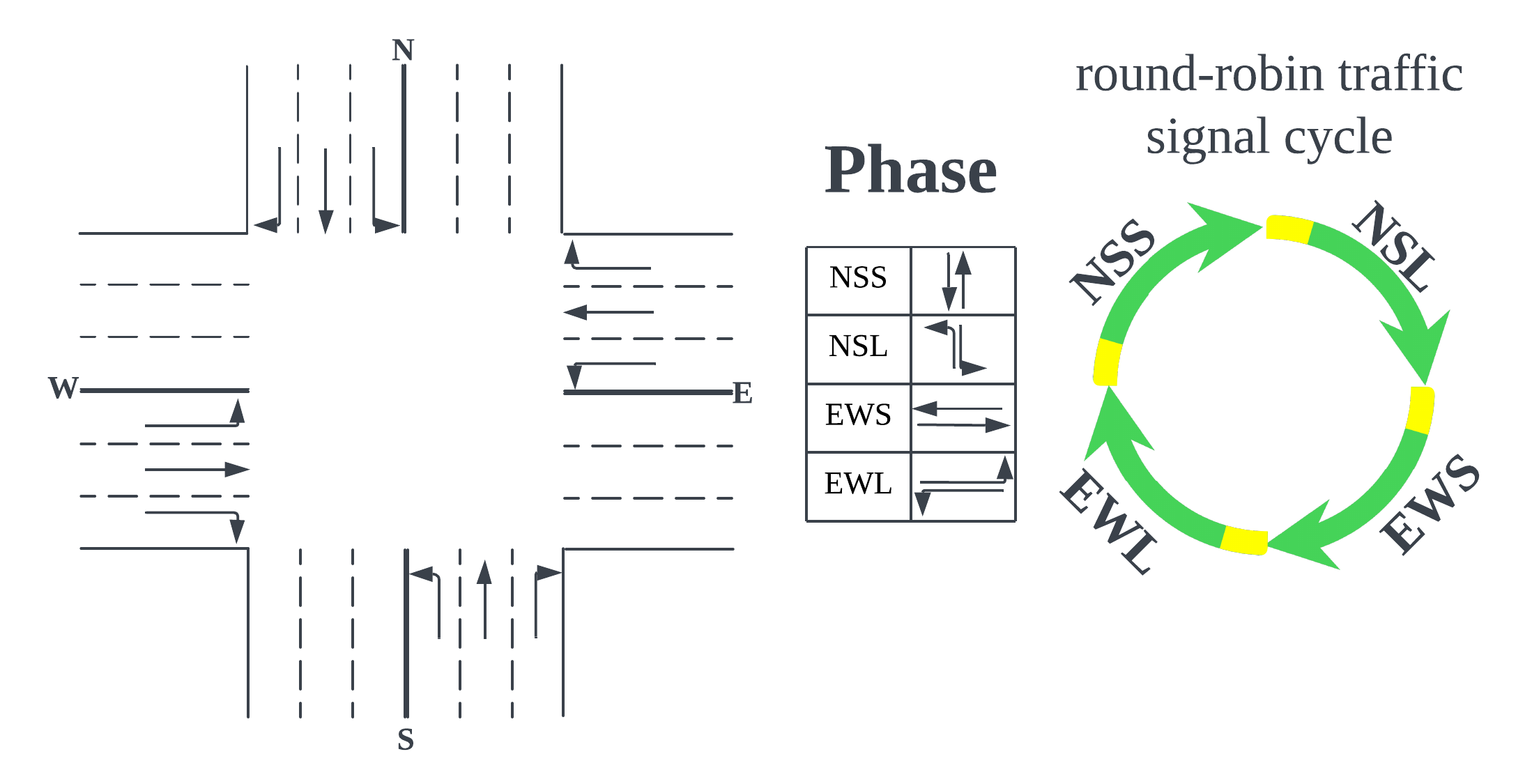}
    \caption{Illustration of an isolated intersection, phase configuration and round-robin traffic signal cycle.}
    \label{fig: traffic network}
\end{figure}
A traffic network is defined as a directed graph $\mathcal{G}=(\mathcal{V},\mathcal{E})$ where $v \in \mathcal{V}$ denotes an intersection and $e_{vu}=(v,u)\in  \mathcal{E}$ denote the road between two intersections as illustrated in Fig.\ref{fig: traffic network}. We consider an intersection with four incoming roads, oriented toward the north (N), south (S), east (E), and west (W). Each road consists of three lanes, with each lane designated for one of the following movements: straight, left turn, or right turn. The combination of these movements defines a traffic signal phase. In this study, we adopt a commonly used four-phase configuration: north–south straight (NSS), north–south left turn (NSL), east–west straight (EWS), and east–west left turn (EWL).

Round Robin traffic signal control is one of the simplest and most widely adopted scheduling strategies in real-world traffic systems\cite{webster1958traffic,koonce2008traffic,little1981maxband}. In this method, signal phases are activated in a fixed, cyclic order, ensuring that each phase receives service in turn. The duration of a complete signal cycle is denoted as $D_{total}$. The objective is to allocate time to each signal phase in a manner that enhances traffic efficiency, subject to the constraint that $D_{NS,\text{straight}}+D_{NS,\text{left}}+D_{EW,\text{straight}}+D_{EW,\text{left}}=D_{total}$
.

\subsection{Deep Deterministic Policy Gradient}
Among the various DRL algorithms applied to traffic signal control, Deep Deterministic Policy Gradient (DDPG) has garnered attention for its effectiveness in handling continuous action spaces\cite{lillicrap2015continuous}. DDPG is an off-policy, actor-critic algorithm that learns a deterministic policy by leveraging the deterministic policy gradient theorem. It employs two neural networks: an actor network $\pi$ with the parameter $\phi$ that maps states $s$ to deterministic actions $a$, and a critic network $Q$ with the parameter $\theta$ that evaluates the expected discounted accumulated rewards of state-action pairs.
The objective is to maximize the expected return 
\begin{align}
    J&=\mathbb{E}_{s \sim \mathcal{D}}\left[ Q(s, a|\theta)  \right]\\
    &=\mathbb{E}_{s \sim \mathcal{D},a\sim\pi(s)}\left[\sum_{t=0}^T\gamma^tr_t\big|s,a\right]
\end{align}
Then, the actor network is updated using the gradient:
\begin{equation}
    \label{eq: actor loss}
    \nabla_{\phi} J \approx \mathbb{E}_{s \sim \mathcal{D}} \left[ \nabla_a Q(s, a|\theta) \big|_{a=\pi(s)} \nabla_{\phi} \pi(s|\phi) \right]
\end{equation}
Here, $\mathcal{D}$ denotes the replay buffer from which state samples are drawn. The critic network is updated by minimizing the loss between the predicted Q-value and the target value:
\begin{equation}
    \label{eq: critic value}
    L(\theta) = \mathbb{E}_{(s, a, r, s') \sim \mathcal{D}} \left[ \left( Q(s, a|\theta) - y \right)^2 \right]
\end{equation}
where the target $y$ is computed as:
\begin{equation}
    y = r + \gamma Q(s', \pi(s'|\phi)|\theta)
\end{equation}
DDPG’s ability to learn continuous control policies makes it particularly suitable for traffic signal environments that require fine-grained adjustments to phase durations or signal timings. This feature allows for a more nuanced approach than discrete action-based methods, bridging the gap between DRL-based control algorithms and real-world traffic management practices.
\section{Methodology}
\begin{figure*}
    \centering
    \includegraphics[width=\textwidth]{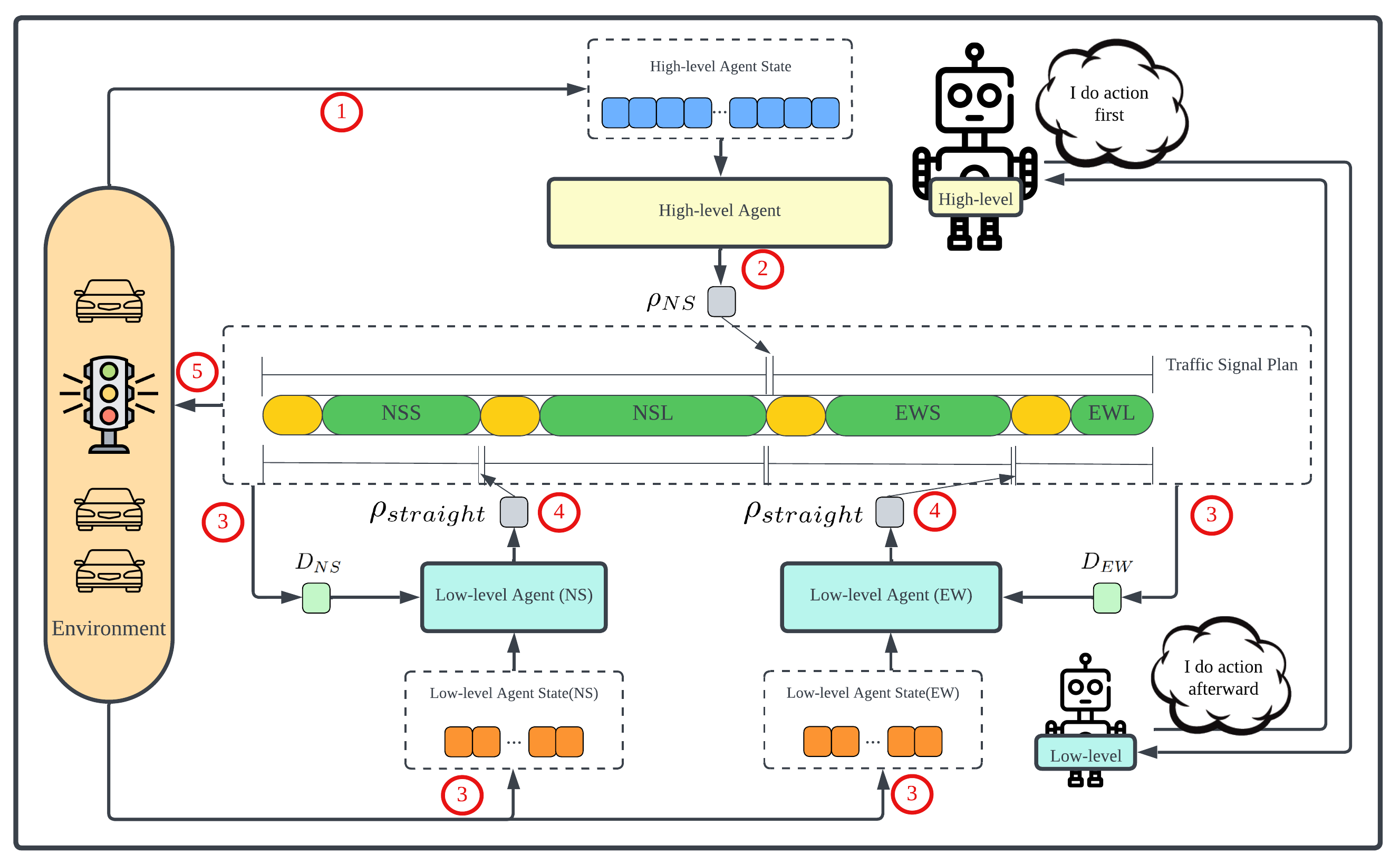}
    \caption{Proposed framework: The high-level agent first retrieves its observation from the environment and then decides the proportion of duration for NS direction. Next, the exact durations of NS direction and EW direction are calculated. Two durations, along with the observations of low-level agents, are further passed to low-level agents. Low-level agents then decide the proportion of duration for straight movement in all directions. Finally, the durations for all phases are calculated and passed to environment.}
    \label{fig:framework}
\end{figure*}

In this section, we describe the framework of our methods and state the formulation of DRL agents. The framework is illustrated in Fig. \ref{fig:framework}. In this framework, two agents are deployed. The high-level agent first observes the traffic state at a given intersection and allocates the total cycle duration between the north–south (NS) and east–west (EW) directions. Subsequently, the low-level agent observes the traffic state for a specific direction, along with the corresponding allocated duration, and further distributes the time between two movements: straight and left turn. Finally, the generated traffic signal plan is applied to the environment, and the phase durations are planned for the next cycle. To prevent any phase from having an excessively short duration, we impose a minimum phase duration constraint, denoted as $D_{min}$.
\subsection{RL Formulation for High-level Agent}
The high-level agent is implemented using DDPG with actor parameters $\phi_h$ and critic parameters $\theta_h$.
\paragraph{State} In the proposed framework, the state representation for the high-level agent focuses on capturing the essential traffic conditions at each intersection. Specifically, the state consists of two key metrics for all incoming lanes: wave and queue length. The wave is defined as the total number of vehicles present on a lane, reflecting the overall traffic volume approaching the intersection. The queue length, on the other hand, measures the number of vehicles that are stationary or moving slowly near the stop line, representing the severity of congestion. Formally, let $\mathcal{L}^{in}$ denote the set of incoming lanes at an intersection. For each lane $m\in \mathcal{L}^{in}$, the high-level state includes $wave_m$ and $queue_m$. The complete state vector $s_h$ for a high-level agent is then given by:
\begin{equation}
    s_h = \{ \text{wave}_m, \text{queue}_m\mid m \in \mathcal{L}^{\text{in}} \}
\end{equation}

\paragraph{Action} The action space for the high-level agent is designed to control the allocation of green time between major directional movements at the intersection. Specifically, the action output is the proportion of the cycle duration assigned to the North-South (NS) traffic direction. Note that since each phase must satisfy the minimum duration constraint $D_{min}$, the action selected by the agent allocates the remaining available time, after accounting for these minimum durations, between the NS and EW directions.
Formally, let $a_h \in [-1,1]$ denote the action selected by the high-level agent, representing the score for NS direction. Then, $a$ is rescaled to $\rho_{NS}\in[0,1]$, which indicates the proportion of duration allocated to the NS phases.
while $1-\rho_{NS}$ corresponds to the proportion allocated to the East-West (EW) phases. 
Given a predefined total cycle length $D_{total}$, the green time durations for the NS and EW directions, denoted $D_{NS}$ and $D_{EW}$ respectively, are computed as:
\begin{align*}
    D_{\text{NS}} &= 2D_{min}+\rho_{NS} \times (D_{total}-4D_{min})\\
    D_{\text{EW}} &= 2D_{min}+(1 - \rho_{NS}) \times(D_{total}-4D_{min})
\end{align*}

\paragraph{Reward} The reward function for the high-level agent is designed to directly encourage the reduction of traffic congestion at the intersection. At each decision step, the agent receives a reward equal to the negative sum of the number of vehicles on all incoming lanes.  The reward $r_h$ at time $t$ is defined as: 
\begin{equation}
    r_{h,t} = -\sum_{m \in \mathcal{L}^{\text{in}}} queue_m
\end{equation}

\subsection{RL Formulation for Low-level Agent on NS direction}
Unlike the high-level agent, which uses traffic states from all incoming lanes to allocate green time between major directions, the low-level agent operates within each major direction to further divide green time between straight and left-turn movements. We first introduce the formulation of the low-level agent in the NS direction. The low-level agent is also implemented using DDPG with actor parameters $\phi_{l}$ and critic parameters $\theta_l$.
\paragraph{State} Specifically, for a given direction (e.g., North-South), the low-level agent observes the traffic state of the lanes associated with straight and left-turn movements in that direction.
Formally,
\begin{equation}
     s_{l,NS} = \{ \text{wave}_m, \text{queue}_m\mid m \in \mathcal{L}^{\text{in}}_{NS}\}\cup \{D_{NS}\}
\end{equation}
where $\mathcal{L}^{\text{in}}_{NS}$ denotes the set of incoming lanes corresponding to the NS direction 
\paragraph{Action}   Based on these observations, it outputs a proportion to split the green time between the two types of movements. 
Formally, let $a_{l,NS}\in[-1,1]$ denote the action selected by the low-level agent, which indicates the score for the Straight Phase. Then, $a^{NS}_l$ is rescaled to $\rho_{NS,straight}\in[0,1]$, which represents the proportion of the direction's green time allocated to the straight movement, and $1-\rho_{NS,straight}$ corresponds to the proportion allocated to the left-turn movement.
Given the green duration on NS direction $D_{NS}$ then the actual green durations for straight ($D_{NS,straight}$) and left-turn ($D_{NS,left}$) are given by:
\begin{align}
    D_{NS,\text{straight}} &= D_{min}+\rho_{NS,straight} \times (D_{NS}-2D_{min})\\
    D_{NS,\text{left}} &= D_{min}+(1 - \rho_{NS,straight}) \times (D_{NS}-2D_{min})
\end{align}
\paragraph{Reward} The reward function for the low-level agent is designed to minimize congestion specifically within its corresponding direction. At each decision step, the low-level agent receives a reward equal to the negative sum of the queue lengths on the straight and left-turn lanes of the assigned direction. The reward $r_{l}$ at time $t$ is then defined as:
\begin{equation}
    r_{l,NS} = -\sum_{m\in \mathcal{L}^{in}_{NS}} queue_m
\end{equation}

\subsection{RL Formulation for Low-level Agent on EW direction}
The formulation for the low-level agent in the EW direction is similar to that of the NS direction, except that it utilizes the traffic state of the EW direction and its action allocates the duration between straight and left-turn movements for the EW direction.
\paragraph{State} 
The state for the low-level agent on the EW direction is
\begin{equation}
     s_{l,EW} = \{ \text{wave}_m, \text{queue}_m\mid m \in \mathcal{L}^{\text{in}}_{EW}\}\cup \{D_{EW}\}
\end{equation}
where $\mathcal{L}^{\text{in}}_{EW}$ denotes the set of incoming lanes corresponding to the NS direction 
\paragraph{Action}  Similar to the action of the low-level agent on the NS direction, let $a_{l,EW}\in[-1,1]$ denote the action selected by the low-level agent on EW direction, which indicates the score for the Straight Phase. Then, $a_{l,EW}$ is rescaled to $\rho_{EW,straight}\in[0,1]$, and the actual green durations for straight ($D_{EW,\text{straight}}$) and left-turn ($D_{EW,\text{left}}$) are given by:
\begin{align}
    D_{EW,\text{straight}} &= D_{min}+\rho_{EW,straight} \times (D_{EW}-2D_{min})\\
    D_{EW,\text{left}} &= D_{min}+(1 - \rho_{EW,straight}) \times (D_{EW}-2D_{min})
\end{align}
\paragraph{Reward} The reward $r_{l,EW}$ at time $t$ is then defined as:
\begin{equation}
    r_{l,EW} = -\sum_{m\in \mathcal{L}^{in}_{EW}} queue_m
\end{equation}
\subsection{Training Algorithm}
To enhance the generalization ability and learning efficiency of the low-level agents, we adopt a parameter-sharing strategy across directions. Specifically, the low-level agents controlling the North-South (NS) and East-West (EW) directions share the same set of network parameters. This design choice leverages the structural similarities between directional traffic patterns and reduces the number of parameters to be trained, leading to faster convergence and better scalability. Furthermore, when extending the framework to a multi-intersection network, we adopt the Centralized Learning and Decentralized Execution (CLDE) paradigm. Under CLDE, all agents are trained in a centralized manner by sharing experience across intersections, while during execution, each agent operates independently based on its own local observations. This approach enables efficient learning by aggregating diverse traffic experiences during training, while maintaining decentralized and scalable control during real-time deployment.
The training procedure is listed in Algorithm \ref{alg:pipeline}.

Both the high-level and low-level agents use fully connected multilayer perceptron (MLP) networks to approximate their policy and value functions. The critic networks are designed with three hidden layers of sizes 300, 200, and 200 neurons, respectively. The actor networks are slightly smaller, consisting of hidden layers with 200, 200, and 100 neurons. During training, to encourage exploration, the action generated by the actor network is perturbed by adding noise sampled from a Gaussian distribution. Specifically, the noise follows $\mathcal{N}(0,0.1)$, where the mean is 0 and the standard deviation is 0.1. This helps the agent to sufficiently explore the action space and avoid premature convergence to suboptimal policies. 

Meanwhile, the traffic signal control simulator used in our experiments only supports discrete, second-scale timing, whereas our model outputs cycle durations in continuous (float) values. To ensure compatibility, we round the generated durations to the nearest integer while adjusting them so that their sum remains equal to the total cycle duration.
\begin{algorithm}
\caption{Algorithm for Pipeline}\label{alg:pipeline}
\begin{algorithmic}[1]
\State Initialise high-level agent networks $\phi_h$ and $\theta_h$ with target networks $\phi_h^-\leftarrow\phi_h$ and $\theta_h^-\leftarrow\theta_h$
\State Initialise low-level agent networks $\phi_l$ and $\theta_l$ with target networks $\phi_l^-\leftarrow\phi_l$ and $\theta_l^-\leftarrow\theta_l$ 
\State Initialise Replay Memory $\mathcal{D}_h$ and $\mathcal{D}_l$ for high-level agent and low-level agent
\While{$i < episode$}
    \State $s_h,s_{l,NS},s_{l,EW}\leftarrow$ environment reset
    \State $a_h \leftarrow \pi(s_h|\phi_h)+\mathcal{N}(0,0.1)$ 
    \State $\rho_{NS} \leftarrow (a_h+1)/2$
    \State $D_{NS}\leftarrow 2D_{min}+\rho_{NS} \times (D_{total}-4D_{min})$ 
    \State $D_{EW}\leftarrow 2D_{min}+(1 - \rho_{NS}) \times(D_{total}-4D_{min})$ 
    \While{$t<T$}
    
        \State $a_{l,NS} \leftarrow \pi(s_{l,NS},D_{NS}|\phi_l)+\mathcal{N}(0,0.1)$
        \State $\rho_{NS,straight} \leftarrow (a_{l,NS}+1)/2$
        \State $D_{NS,\text{straight}} \leftarrow D_{min}+\rho_{NS,straight} \times (D_{NS}-2D_{min})$
   \State $D_{NS,\text{left}} \leftarrow D_{min}+(1 - \rho_{NS,straight}) \times (D_{NS}-2D_{min})$
        \State $a^{EW}_l \leftarrow \pi(s_{l,EW},D_{EW}|\phi_l)+\mathcal{N}(0,0.1)$
        \State $\rho_{EW,straight} \leftarrow (a_{l,EW}+1)/2$   
        \State $D_{EW,\text{straight}} \leftarrow D_{min}+\rho_{EW,straight} \times (D_{EW}-2D_{min})$
   \State $D_{EW,\text{left}} \leftarrow D_{min}+(1 - \rho_{EW,straight}) \times (D_{EW}-2D_{min})$
        \State $r_h,r_{l,NS},r_{l,EW},s_{h}',s_{l,NS}',s_{l,EW}'\leftarrow$  environment step 
        based 
        on the following plan 
        \Statex \hspace*{3em}$(D_{NS,\text{straight}},D_{NS,\text{left}},D_{EW,\text{straight}},D_{EW,\text{left}})$ 
        \State $a'_h \leftarrow \pi(s'_h|\phi_h)+\mathcal{N}(0,0.1)$ 
    \State $\rho'_{NS} \leftarrow (a_h+1)/2$
    \State $D'_{NS}\leftarrow 2D_{min}+\rho'_{NS} \times (D_{total}-4D_{min})$ 
    \State $D'_{EW}\leftarrow 2D_{min}+(1 - \rho'_{NS}) \times(D_{total}-4D_{min})$ 
        \State store transition$(s_h,a_h,r_h,s_h')$ to $\mathcal{D}_h$
        \State store transition 
        \Statex \hspace*{3em} $(s_{l,NS}\cup D_{NS},a_{l,NS},r_{l,NS},s'_{l,NS}\cup D'_{NS})$ to $\mathcal{D}_l$
        \State store transition 
        \Statex \hspace*{3em}$(s_{l,EW}\cup D_{EW},a_{l,EW},r_{l,EW},s'_{l,EW}\cup D'_{EW})$ to $\mathcal{D}_l$
        \State sample experience batches from $\mathcal{D}_h$ and $\mathcal{D}_l$
        \State update $\phi_h, \theta_h, \phi_l,\theta_l$  by Eq. (\ref{eq: actor loss}) and (\ref{eq: critic value}).
        \State $\theta_h^-\leftarrow (1-\tau)\theta_h^-+\tau \theta_h$
      \State $\theta_l^-\leftarrow (1-\tau)\theta_l^-+\tau \theta_l$
        \State $\phi_h^-\leftarrow (1-\tau)\phi_h^-+\tau \phi_h$
      \State $\phi_l^-\leftarrow (1-\tau)\phi_l^-+\tau \phi_l$
        \State $D_{NS}\leftarrow D_{NS}'$
        \State $D_{EW}\leftarrow D_{EW}'$
        
    \EndWhile

\EndWhile
\end{algorithmic}
\end{algorithm}
\section{Experiments and Discussion}
\subsection{Simulator Setting and Test Scenarios}
\begin{figure}
    \centering
    \subfloat[Jinan]{
    \includegraphics[width=0.33\textwidth]{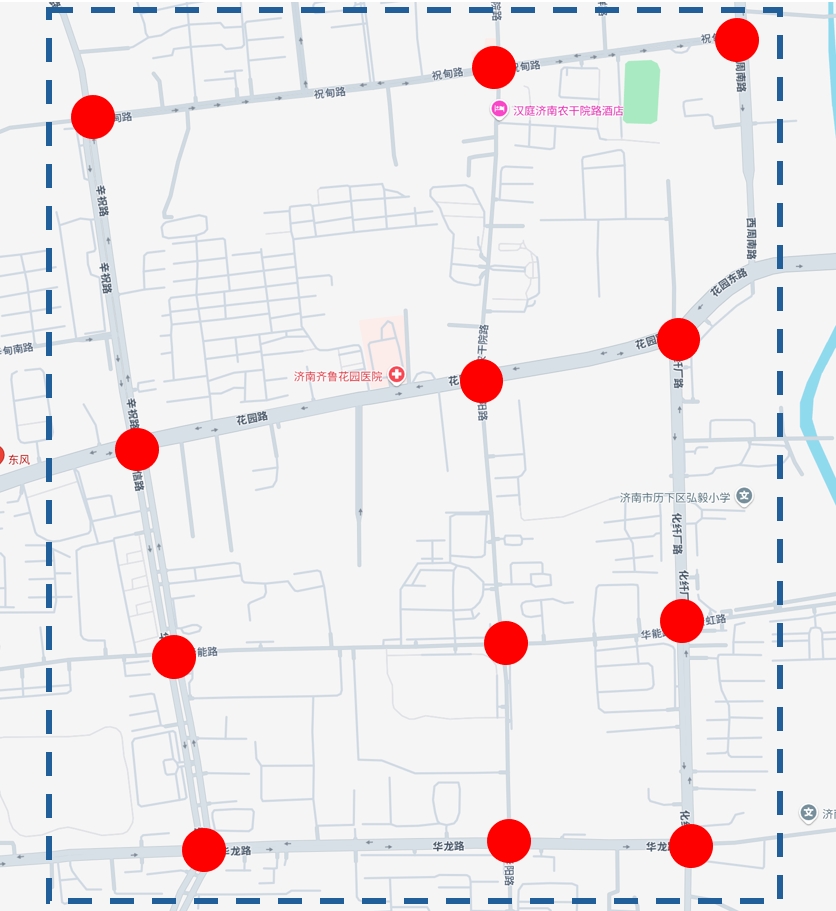}
    }
    
    \subfloat[Hangzhou]{
    \includegraphics[width=0.33\textwidth]{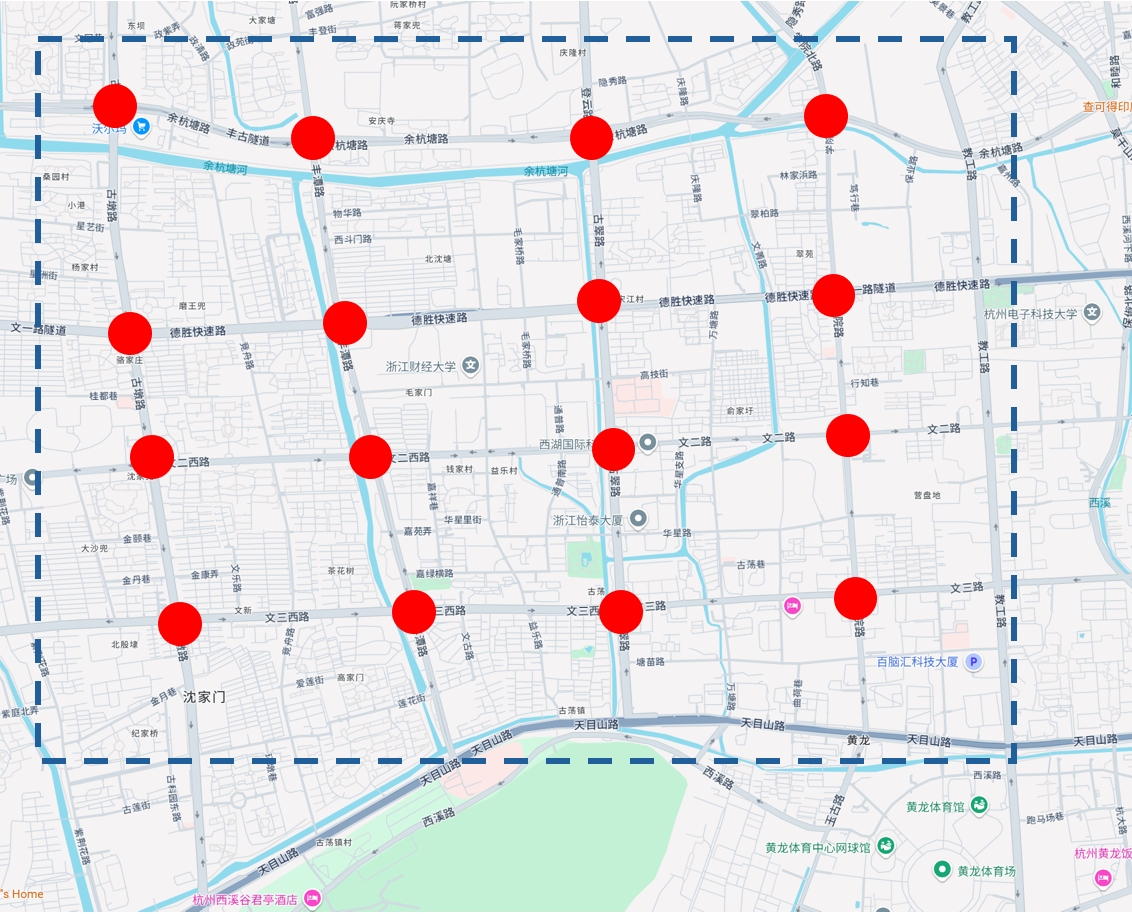}
    }
    \caption{Road Network for Jinan and Hangzhou.}
    \label{fig: roadnet}
\end{figure}
We test the performance of our framework on the open-sourced traffic simulator Cityflow\cite{zhang2019cityflow}.
Our experimental evaluation is conducted on three distinct networks: a real-world network from Jinan, a real-world network from Hangzhou, and a synthetic network designed for controlled testing. The traffic flow data for the Jinan and Hangzhou networks are collected using camera-based vehicle detection systems, providing realistic and heterogeneous traffic patterns. For the Jinan network, we use three different traffic flow profiles, each representing varying demand levels and directional imbalances observed during different times of day. For the Hangzhou network, two distinct traffic flow profiles are utilized to capture diverse traffic conditions. In contrast, the traffic flow for the synthetic network is generated artificially following a Gaussian distribution, allowing for controlled variation and reproducibility in experimental setups. This combination of real-world and synthetic scenarios ensures a comprehensive evaluation of the proposed method across diverse traffic environments. The data for road networks and traffic flows is available open-sourced online\footnote{https://traffic-signal-control.github.io/\#open-datasets}.
\begin{table}[]
    \centering
    \caption{The Configuration of Scenarios}
    \begin{tabular}{|c|c|c|}
         \hline
         Scenario &Approach Length & Total Trajectories\\

         \hline
        Jinan(Flow1)& 400m(EW),800m(NS)&6295\\ 
         \hline
         Jinan(Flow2)& 400m(EW),800m(NS)&4365\\ 
         \hline
         Jinan(Flow3)& 400m(EW),800m(NS)&5494\\ 
         \hline
         Hangzhou(Flat)& 800m(EW),600m(NS)&2983\\ 
         \hline
         Hangzhou(Peak) &800m(EW),600m(NS)&6538\\
        \hline
        Synthetic & 300m(EW),300m(NS)&11231\\
        \hline
    \end{tabular}

    \label{tab: traffic info}
\end{table}
The simulation time for each experiment is set to 3600 seconds (1 hour) to capture realistic traffic dynamics over an extended period. In our method, the agent makes decisions every 60 seconds, where each decision corresponds to adjusting the parameters for a full traffic signal cycle. During each 60-second cycle, the signal phases are executed according to the durations determined by the high-level and low-level agents. In contrast, baseline models are configured to make decisions every 15 seconds, reflecting a more frequent but less structured control scheme. To ensure safety when transitioning between phases, a 3-second yellow phase is inserted whenever a phase change occurs, allowing vehicles already within the intersection to clear before the next movement begins. This yellow-phase design is critical for preventing conflicts and ensuring safe operation during phase switching.
\begin{table}[htp]
    \centering
        \caption{Hyperparameter Configuration}
    \begin{tabular}{|c|c|}
    \hline
         Name & Value  \\
        \hline
        $D_{min}$ &5s\\\hline
        $D_{total}$ & 60s\\\hline
        $\gamma$ &0.9  \\ \hline
         Replay Buffer Size&100000\\\hline
         Episode Number& 2000\\\hline
         Network optimizer&Adam\\\hline
         Actor Learning Rate&1e-4 \\\hline
         Critic Learning Rate&1e-3 \\\hline
         Batch Size &128 \\\hline         
         Soft update rate&1e-4\\\hline                         
    \end{tabular}

    \label{tab: hyperparameter}
\end{table}
\begin{figure*}[htp]
    \centering
    
    \subfloat[Jinan(Flow1)]{
    \includegraphics[width=0.3\textwidth]{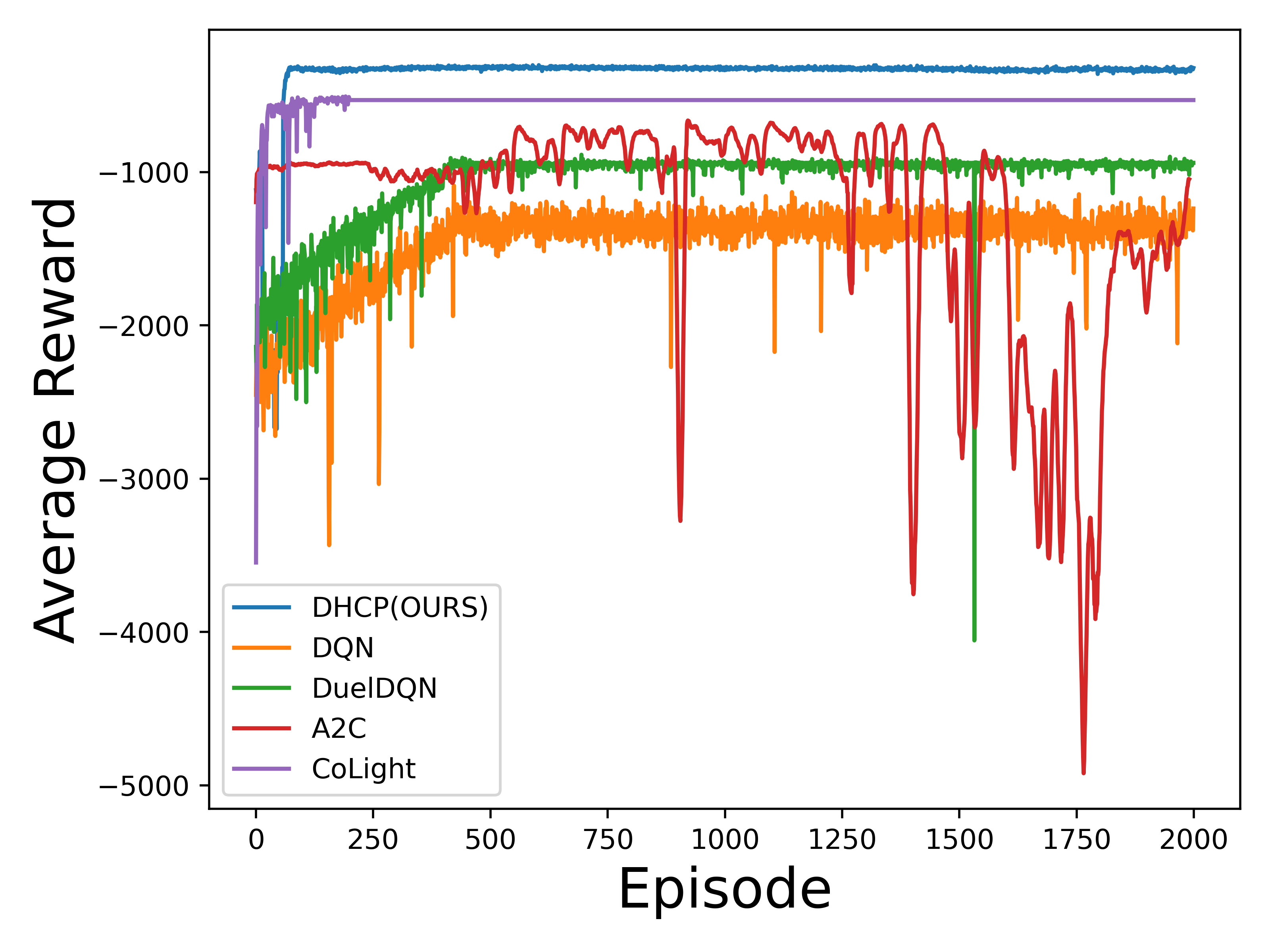}
    }
        \subfloat[Jinan(Flow2)]{
    \includegraphics[width=0.3\textwidth]{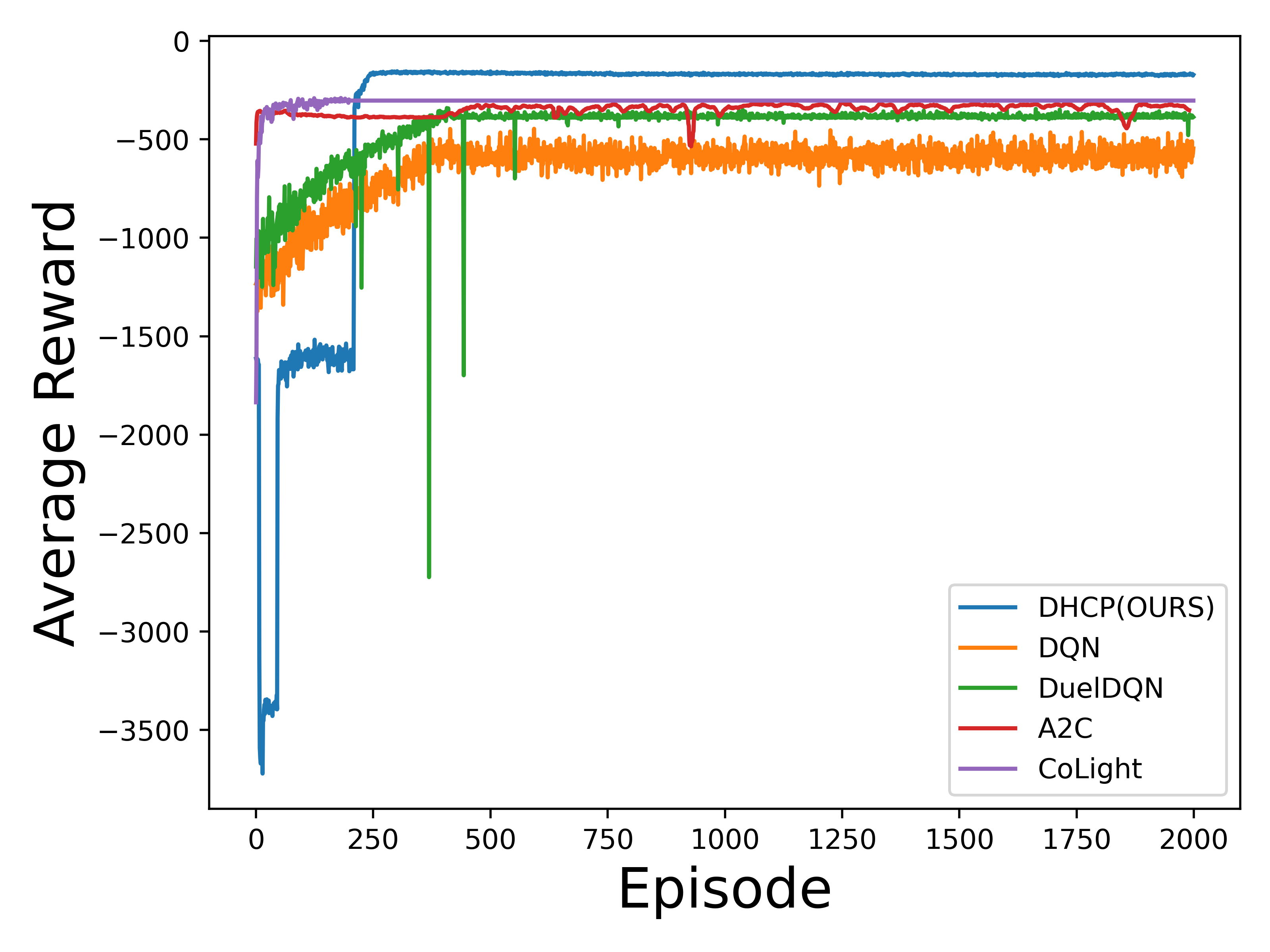}
    }
        \subfloat[Jinan(Flow3)]{
    \includegraphics[width=0.3\textwidth]{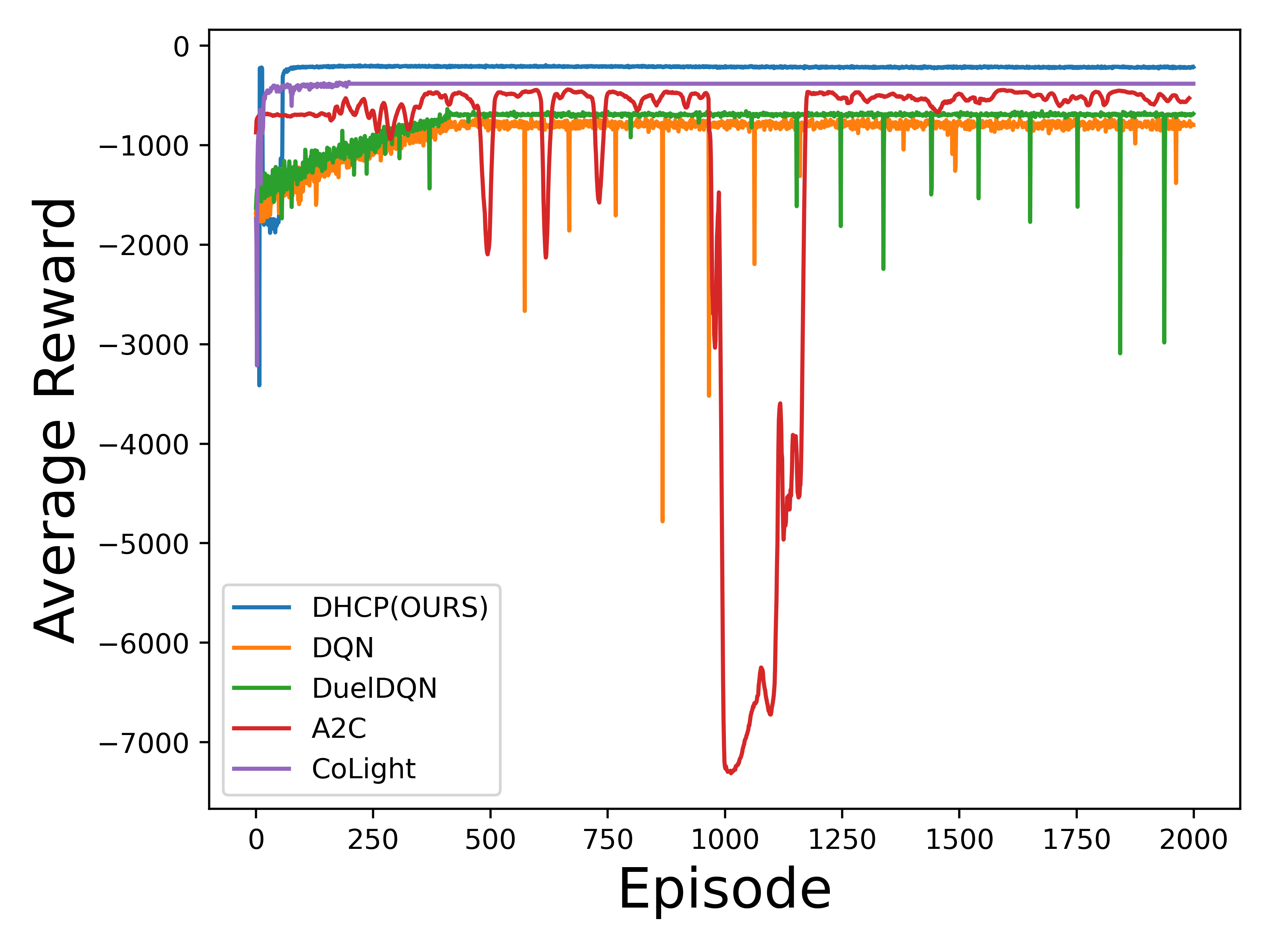}
    }
    
    \subfloat[Hangzhou(Flat)]{
    \includegraphics[width=0.3\textwidth]{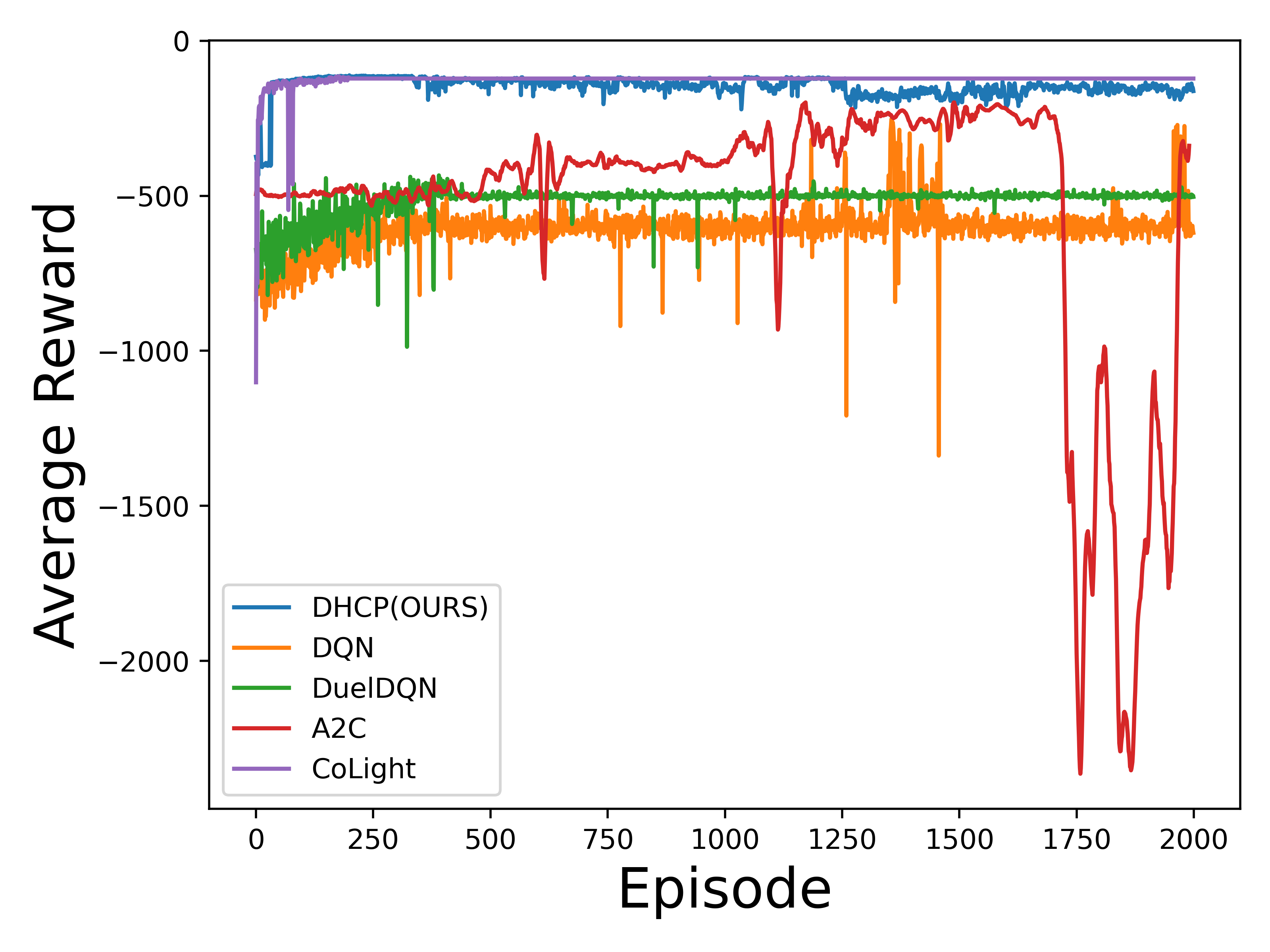}
    }
    \subfloat[Hangzhou(Peak)]{
    \includegraphics[width=0.3\textwidth]{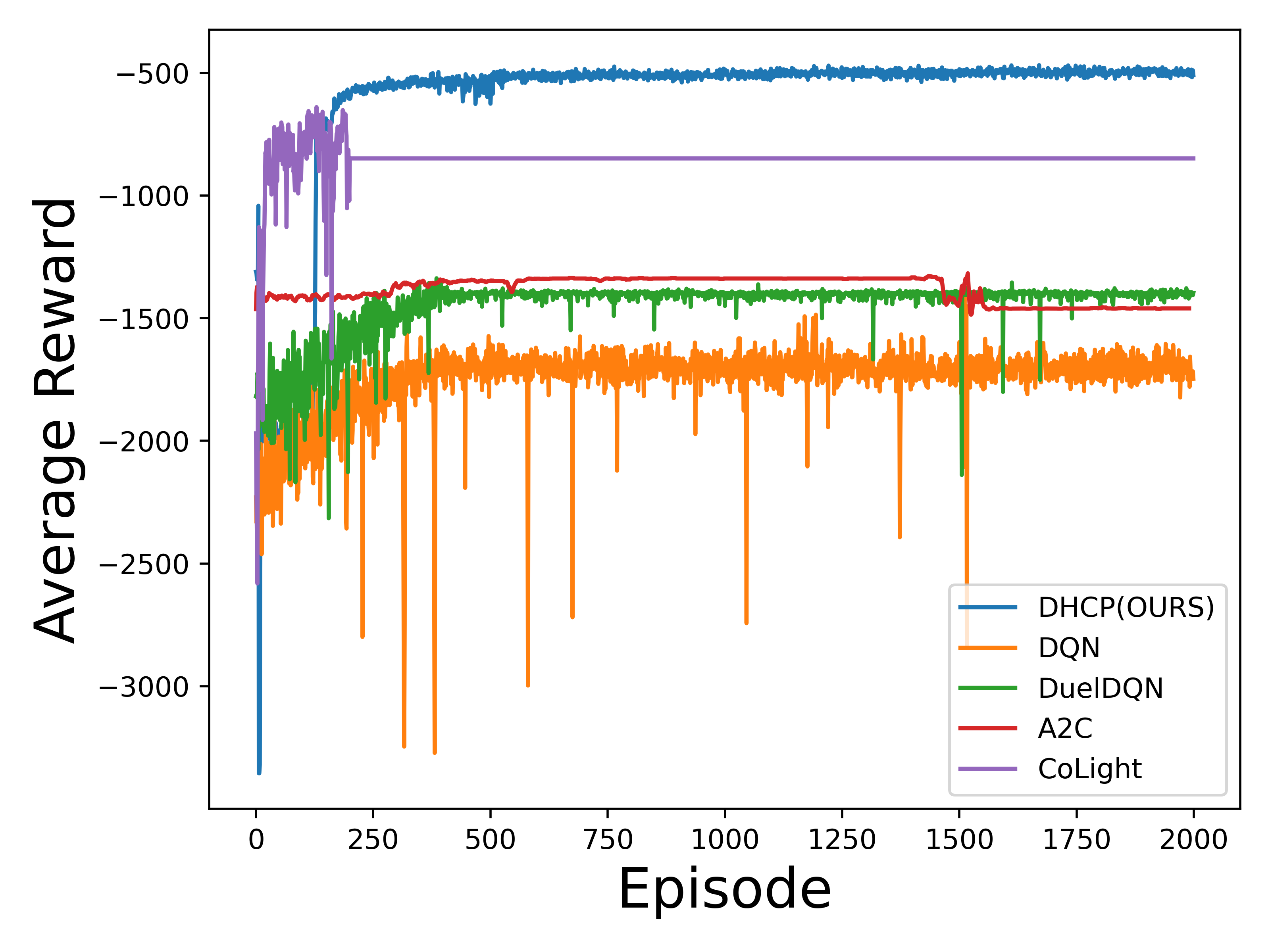}
    }
    \subfloat[Synthetic]{
    \includegraphics[width=0.3\textwidth]{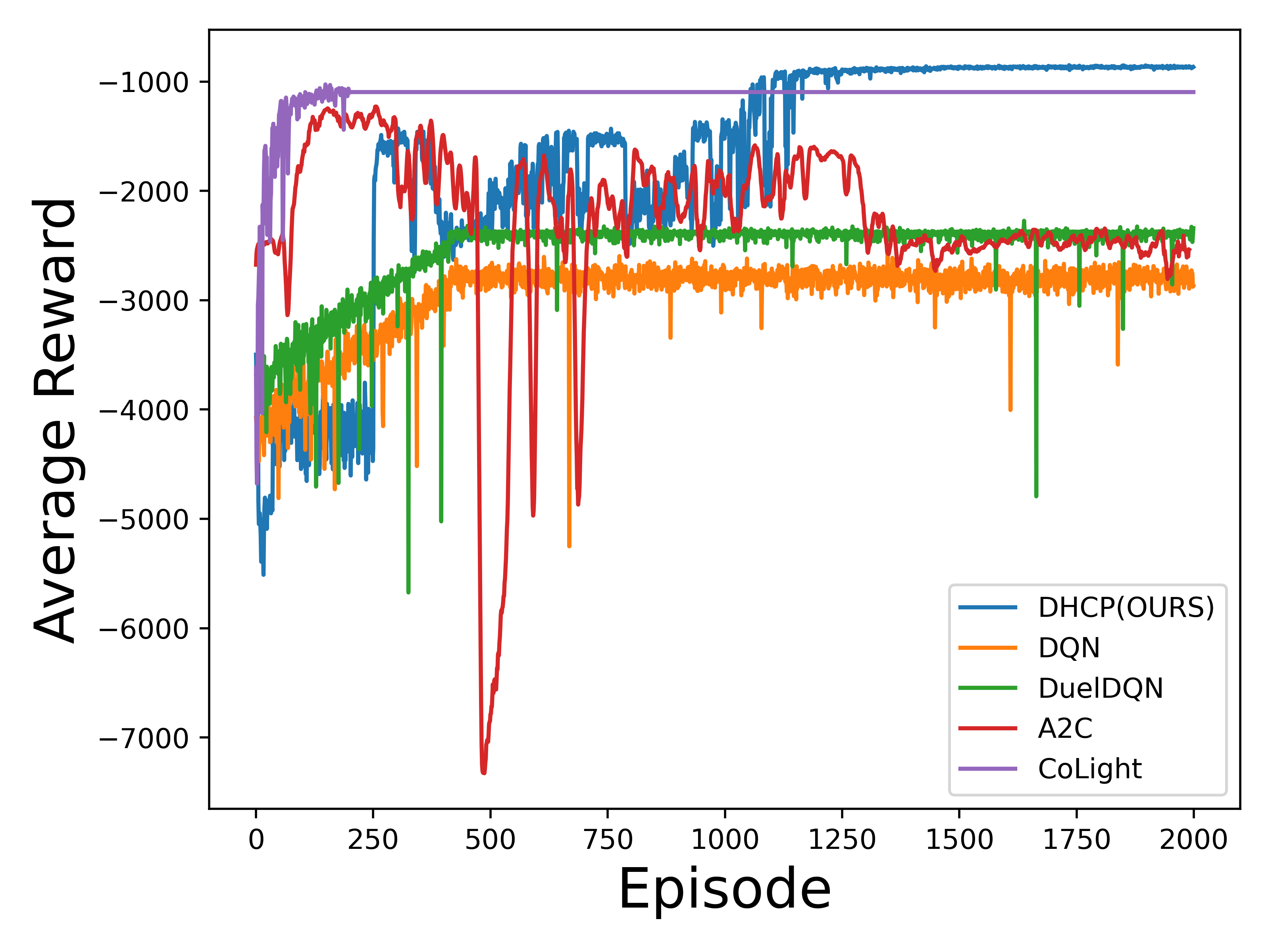}
    }

    \caption{Average Episode Reward Curve During Training.}
    \label{fig: episode reward curve}
\end{figure*}
\subsection{Baseline}
\begin{itemize}
    \item Fixed-time traffic signal control: This method is a traditional signal control method where the phase durations and sequences are pre-defined and remain constant, regardless of real-time traffic conditions.
    \item Self-Organizing Traffic Lights (SOLT)\cite{gershenson2004self}: SOLT is an adaptive traffic signal control approach that dynamically adjusts signal phases based on real-time traffic demand, optimizing traffic flow without predefined schedules.
    \item MaxPressure\cite{varaiya2013max}: MaxPressure is an adaptive traffic signal control strategy that selects phases to maximize traffic flow by minimizing queue lengths and congestion at intersections in real-time.
    \item DQN\cite{mnih2015human}: DQN is a classic deep reinforcement method that applies one evaluation network and one target network during training.
    \item Dueling-DQN\cite{wang2016dueling}: Dueling-DQN is an extension DRL network of DQN with advantage value estimation to stabilize the training process.
    \item Advantage Actor-Critic (A2C): A2C is a reinforcement learning algorithm that combines both an actor and a critic to optimize the policy, using the advantage function to reduce variance in policy updates and improve learning efficiency.
    \item Colight\cite{wei2019colight}: CoLight is a state-of-the-art DRL model that applies a graph attention layer to compute the correlation between an intersection and its neighborhoods.

\end{itemize}

\subsection{Results}

The first evaluation metric is the average summed reward across all intersections within each episode for DRL-based methods. This metric reflects the overall learning performance and convergence behavior of the agents during training. In our model, rewards are collected every 60 seconds, corresponding to one full traffic signal cycle, whereas for DRL baselines, rewards are collected every 15 seconds. To ensure a consistent scale for comparison, we aggregate the baseline rewards by taking the average over four consecutive 15-second intervals, thereby aligning the reward evaluation with the 60-second cycle used in our model. From the results across the six test scenarios (Fig. \ref{fig: episode reward curve}), our proposed model consistently achieves the best performance compared to all baselines. Both DQN and Dueling-DQN show gradual improvements during training; however, their final performance plateaus and remains limited after convergence. Notably, Dueling-DQN demonstrates greater stability than DQN, exhibiting less fluctuation in reward trends. In contrast, A2C presents higher instability due to its reliance on Monte Carlo sampling, and its performance even deteriorates during the later stages of training. Our model, DHCP, demonstrates superior performance after reaching convergence. However, it is important to note that the performance metric of our model is relatively poor during the early stages of training. This can be attributed to the hierarchical structure of our approach, where two DDPG agents (high-level and low-level) must learn to cooperate effectively. In the early phase, both agents need to adapt to each other: the high-level agent must learn how to allocate duration efficiently for the low-level agent, while the low-level agent needs to optimize its usage of the allocated duration to achieve the best local objective. This mutual adaptation process results in the initial performance dip before the model stabilizes and achieves optimal results. After sufficient exploration, our model converges rapidly in most scenarios, demonstrating quick adaptation and learning. In the synthetic scenario, the convergence is slower, and the convergence curve exhibits a pattern of improvement, followed by a decrease, and then an eventual increase. This fluctuation suggests that the two agents are undergoing an adaptation process with each other. This mutual adaptation is likely helping the model to escape from local optima, ultimately leading to better performance after sufficient exploration and fine-tuning.

\begin{table*}[htb]
\renewcommand{\arraystretch}{2}
\centering
\caption{Average Travel Time in Testing Scenarios}
\begin{tabular}{l|llllll}
\hline
Model & Jinan(Flow1)& Jinan(Flow2) &Jinan(Flow3)&Hangzhou(Flat)  & Hangzhou(Peak)  & Synthetic   \\ \hline
Fixed  Time& 713.72&695.10& 660.93&699.74&972.16& 847.05\\
SOLT &559.35&304.04&301.72&357.65&440.48&301.68\\
Max Pressure& 297.08&288.02&300.96&332.76&446.94&253.18\\
DQN & 368.75$\pm$0.82&303.03$\pm$0.93&330.92$\pm$2.91&413.84$\pm$5.14&599.57$\pm$6.32&503.40$\pm$2.47\\
Dueling-DQN& 353.54$\pm$0.76&297.13$\pm$1.04&322.82$\pm$0.91&440.48$\pm$4.05&598.92$\pm$ 4.68 & 491.83$\pm$2.67\\
A2C&312.84$\pm$5.81&298.59$\pm$0.68&287.71$\pm$0.84&353.53$\pm$12.87&594.26$\pm$1.54&448.62$\pm$4.63\\
Colight& 296.99$\pm$0.94&290.09$\pm$0.54&282.72$\pm$0.75&331.34$\pm$0.90&448.74$\pm$4.61&240.91$\pm$2.07\\
\hline
DHCP & 292.15$\pm$0.49 & 287.27$\pm$0.46&280.27$\pm$0.26 &327.30$\pm$0.38 &394.37$\pm$0.78 &221.74$\pm$0.49\\\hline
\end{tabular}
\label{tab: general results}
\end{table*}
The second evaluation metric is average travel time, which offers a comprehensive measure of the model's effectiveness in optimizing traffic flow across all intersections. This metric is crucial as it directly reflects the efficiency of the traffic signal control system in reducing delays for vehicles. A lower average travel time indicates that the model is effectively managing the traffic signal phases, leading to smoother traffic flow and reduced congestion. As shown in the results (Table \ref{tab: general results}), the average travel time achieved by our model is the shortest across all test scenarios. Our method significantly outperforms the baselines, especially the fixed-time control under the same cycle length, demonstrating much better adaptability to varying traffic conditions. These results highlight the effectiveness of our hierarchical DRL approach in improving overall traffic efficiency.

\section{Conclusion}
In this paper, we propose a novel deep reinforcement learning (DRL) model that hierarchically allocates traffic signal cycle durations. A high-level agent first determines the split of the total cycle time between the North-South (NS) and East-West (EW) directions. Subsequently, a low-level agent further divides the allocated time between straight and left-turn movements within each major direction. Both agents are implemented using the DDPG algorithm, which generates continuous actions well-suited to representing the proportion of cycle duration allocations. Experimental results demonstrate the superiority of our proposed model across two key metrics, showcasing its effectiveness in both learning performance and average travel time. 

One limitation of our model is that it assumes a fixed total duration for every traffic signal cycle and does not explicitly consider coordination between neighboring intersections. Meanwhile, we only consider four-way intersections in this paper. In future work, we will focus on the coordination of traffic signal cycles across multiple intersections and extend our model to more complex networks with heterogeneous intersections.

\bibliographystyle{IEEEtran}
\bibliography{reference}

\begin{thebibliography}{10}
\providecommand{\url}[1]{#1}
\csname url@samestyle\endcsname
\providecommand{\newblock}{\relax}
\providecommand{\bibinfo}[2]{#2}
\providecommand{\BIBentrySTDinterwordspacing}{\spaceskip=0pt\relax}
\providecommand{\BIBentryALTinterwordstretchfactor}{4}
\providecommand{\BIBentryALTinterwordspacing}{\spaceskip=\fontdimen2\font plus
\BIBentryALTinterwordstretchfactor\fontdimen3\font minus \fontdimen4\font\relax}
\providecommand{\BIBforeignlanguage}[2]{{%
\expandafter\ifx\csname l@#1\endcsname\relax
\typeout{** WARNING: IEEEtran.bst: No hyphenation pattern has been}%
\typeout{** loaded for the language `#1'. Using the pattern for}%
\typeout{** the default language instead.}%
\else
\language=\csname l@#1\endcsname
\fi
#2}}
\providecommand{\BIBdecl}{\relax}
\BIBdecl

\bibitem{hussain2023investigating}
Z.~Hussain, M.~Kaleem~Khan, and Z.~Xia, ``Investigating the role of green transport, environmental taxes and expenditures in mitigating the transport co2 emissions,'' \emph{Transportation Letters}, vol.~15, no.~5, pp. 439--449, 2023.

\bibitem{geng2016environmental}
Z.~Geng, Q.~Chen, Q.~Xia, D.~S. Kirschen, and C.~Kang, ``Environmental generation scheduling considering air pollution control technologies and weather effects,'' \emph{IEEE Transactions on Power Systems}, vol.~32, no.~1, pp. 127--136, 2016.

\bibitem{brilon2013experiences}
W.~Brilon and T.~Wietholt, ``Experiences with adaptive signal control in germany,'' \emph{Transportation research record}, vol. 2356, no.~1, pp. 9--16, 2013.

\bibitem{wei2019survey}
H.~Wei, G.~Zheng, V.~Gayah, and Z.~Li, ``A survey on traffic signal control methods,'' \emph{arXiv preprint arXiv:1904.08117}, 2019.

\bibitem{haydari2020deep}
A.~Haydari and Y.~Y{\i}lmaz, ``Deep reinforcement learning for intelligent transportation systems: A survey,'' \emph{IEEE Transactions on Intelligent Transportation Systems}, vol.~23, no.~1, pp. 11--32, 2020.

\bibitem{wu2021efficient}
Q.~Wu, L.~Zhang, J.~Shen, L.~L{\"u}, B.~Du, and J.~Wu, ``Efficient pressure: Improving efficiency for signalized intersections,'' \emph{arXiv preprint arXiv:2112.02336}, 2021.

\bibitem{Liu2025MATLIT}
B.~Liu, K.~Su, E.~Wang, W.~Han, L.~Wu, J.~Wang, and C.~Qiao, ``Matlit: Mat-based cooperative reinforcement learning for urban traffic signal control,'' \emph{IEEE Transactions on Intelligent Transportation Systems}, pp. 1--15, 2025.

\bibitem{gu2024large}
H.~Gu, S.~Wang, X.~Ma, D.~Jia, G.~Mao, E.~G. Lim, and C.~P.~R. Wong, ``Large-scale traffic signal control using constrained network partition and adaptive deep reinforcement learning,'' \emph{IEEE Transactions on Intelligent Transportation Systems}, vol.~25, no.~7, pp. 7619--7632, 2024.

\bibitem{hankang2025Communication}
H.~Gu, S.~Wang, D.~Jia, Y.~Zhang, Y.~Luo, G.~Mao, J.~Wang, and E.~G. Lim, ``Communication strategy on macro-and-micro traffic state in cooperative deep reinforcement learning for regional traffic signal control,'' \emph{IEEE Transactions on Intelligent Transportation Systems}, pp. 1--14, 2025.

\bibitem{jing2022HALight}
J.~Zeng, J.~Xin, Y.~Cong, J.~Zhu, Y.~Zhang, W.~Jiang, and S.~Pu, ``Halight: Hierarchical deep reinforcement learning for cooperative arterial traffic signal control with cycle strategy,'' in \emph{2022 IEEE 25th International Conference on Intelligent Transportation Systems (ITSC)}, 2022, pp. 479--485.

\bibitem{tan1993multi}
M.~Tan, ``Multi-agent reinforcement learning: Independent vs. cooperative agents,'' in \emph{Proceedings of the tenth international conference on machine learning}, 1993, pp. 330--337.

\bibitem{wei2019presslight}
H.~Wei, C.~Chen, G.~Zheng, K.~Wu, V.~Gayah, K.~Xu, and Z.~Li, ``Presslight: Learning max pressure control to coordinate traffic signals in arterial network,'' in \emph{Proceedings of the 25th ACM SIGKDD International Conference on Knowledge Discovery \& Data Mining}, 2019, pp. 1290--1298.

\bibitem{wei2019colight}
H.~Wei, N.~Xu, H.~Zhang, G.~Zheng, X.~Zang, C.~Chen, W.~Zhang, Y.~Zhu, K.~Xu, and Z.~Li, ``Colight: Learning network-level cooperation for traffic signal control,'' in \emph{Proceedings of the 28th ACM International Conference on Information and Knowledge Management}, 2019, pp. 1913--1922.

\bibitem{zhang2023large}
Y.~Zhang, S.~Wang, X.~Ma, W.~Yue, and R.~Jiang, ``Large-scale traffic signal control by a nash deep q-network approach,'' in \emph{2023 IEEE 26th International Conference on Intelligent Transportation Systems (ITSC)}.\hskip 1em plus 0.5em minus 0.4em\relax IEEE, 2023, pp. 4584--4591.

\bibitem{yang2018hierarchical}
Z.~Yang, K.~Merrick, L.~Jin, and H.~A. Abbass, ``Hierarchical deep reinforcement learning for continuous action control,'' \emph{IEEE transactions on neural networks and learning systems}, vol.~29, no.~11, pp. 5174--5184, 2018.

\bibitem{vezhnevets2017feudal}
A.~S. Vezhnevets, S.~Osindero, T.~Schaul, N.~Heess, M.~Jaderberg, D.~Silver, and K.~Kavukcuoglu, ``Feudal networks for hierarchical reinforcement learning,'' in \emph{International conference on machine learning}.\hskip 1em plus 0.5em minus 0.4em\relax PMLR, 2017, pp. 3540--3549.

\bibitem{xu2021hierarchically}
B.~Xu, Y.~Wang, Z.~Wang, H.~Jia, and Z.~Lu, ``Hierarchically and cooperatively learning traffic signal control,'' in \emph{Proceedings of the AAAI conference on artificial intelligence}, vol.~35, no.~1, 2021, pp. 669--677.

\bibitem{hunt1982scoot}
P.~Hunt, D.~Robertson, R.~Bretherton, and M.~C. Royle, ``The scoot on-line traffic signal optimisation technique,'' \emph{Traffic Engineering \& Control}, vol.~23, no.~4, 1982.

\bibitem{wei2018intellilight}
H.~Wei, G.~Zheng, H.~Yao, and Z.~Li, ``Intellilight: A reinforcement learning approach for intelligent traffic light control,'' in \emph{Proceedings of the 24th ACM SIGKDD International Conference on Knowledge Discovery \& Data Mining}, 2018, pp. 2496--2505.

\bibitem{van2016coordinated}
E.~Van~der Pol and F.~A. Oliehoek, ``Coordinated deep reinforcement learners for traffic light control,'' \emph{Proceedings of learning, inference and control of multi-agent systems (at NIPS 2016)}, vol.~1, 2016.

\bibitem{mannion2016experimental}
P.~Mannion, J.~Duggan, and E.~Howley, ``An experimental review of reinforcement learning algorithms for adaptive traffic signal control,'' \emph{Autonomic road transport support systems}, pp. 47--66, 2016.

\bibitem{lillicrap2015continuous}
T.~P. Lillicrap, J.~J. Hunt, A.~Pritzel, N.~Heess, T.~Erez, Y.~Tassa, D.~Silver, and D.~Wierstra, ``Continuous control with deep reinforcement learning,'' \emph{arXiv preprint arXiv:1509.02971}, 2015.

\bibitem{webster1958traffic}
F.~V. Webster, ``Traffic signal settings,'' Tech. Rep., 1958.

\bibitem{koonce2008traffic}
P.~Koonce and L.~Rodegerdts, ``Traffic signal timing manual.'' United States. Federal Highway Administration, Tech. Rep., 2008.

\bibitem{little1981maxband}
J.~D. Little, M.~D. Kelson, and N.~H. Gartner, ``Maxband: A versatile program for setting signals on arteries and triangular networks,'' 1981.

\bibitem{zhang2019cityflow}
H.~Zhang, S.~Feng, C.~Liu, Y.~Ding, Y.~Zhu, Z.~Zhou, W.~Zhang, Y.~Yu, H.~Jin, and Z.~Li, ``Cityflow: A multi-agent reinforcement learning environment for large scale city traffic scenario,'' in \emph{The world wide web conference}, 2019, pp. 3620--3624.

\bibitem{gershenson2004self}
C.~Gershenson, ``Self-organizing traffic lights,'' \emph{arXiv preprint nlin/0411066}, 2004.

\bibitem{varaiya2013max}
P.~Varaiya, ``Max pressure control of a network of signalized intersections,'' \emph{Transportation Research Part C: Emerging Technologies}, vol.~36, pp. 177--195, 2013.

\bibitem{mnih2015human}
V.~Mnih, K.~Kavukcuoglu, D.~Silver, A.~A. Rusu, J.~Veness, M.~G. Bellemare, A.~Graves, M.~Riedmiller, A.~K. Fidjeland, G.~Ostrovski \emph{et~al.}, ``Human-level control through deep reinforcement learning,'' \emph{nature}, vol. 518, no. 7540, pp. 529--533, 2015.

\bibitem{wang2016dueling}
Z.~Wang, T.~Schaul, M.~Hessel, H.~Hasselt, M.~Lanctot, and N.~Freitas, ``Dueling network architectures for deep reinforcement learning,'' in \emph{International conference on machine learning}.\hskip 1em plus 0.5em minus 0.4em\relax PMLR, 2016, pp. 1995--2003.

\end{thebibliography}


\end{document}